\documentclass{article}
\usepackage{ijcai20}

\usepackage{times}

\usepackage{soul}
\usepackage{url}
\usepackage[hidelinks]{hyperref}
\usepackage[utf8]{inputenc}
\usepackage[small]{caption}
\usepackage{graphicx}
\usepackage{mathtools}
\usepackage{amsmath}
\usepackage{multirow}
\usepackage{amssymb}
\usepackage[font=small]{caption}
\usepackage{subcaption}
\usepackage[boxed]{algorithm2e}

\usepackage{booktabs}
\newcommand{\comment}[1]{}
\SetAlCapSkip{1em}

\SetKwInput{KwInput}{Input}
\SetKwInput{KwOutput}{Output}

\DeclareMathOperator*{\argmin}{arg\,min}

\title{Integrating Image Captioning with Rule-based Entity Masking}

\author{
Aditya Mogadala $^1$\footnote{Contact Author}\and
Xiaoyu Shen $^{1,2}$ \and
Dietrich Klakow $^1$\\
\affiliations
$^1$Spoken Language Systems, Saarland Informatics Campus, Saarland University\\
$^2$Max Planck Institute for Informatics\\
\emails
amogadala@lsv.uni-saarland.de,
xshen@mpi-inf.mpg.de,
dietrich.klakow@uni-saarland.de
}

\begin{document}

\maketitle

\begin{abstract}
Given an image, generating its natural language description (i.e., caption) is a well studied problem. Approaches proposed to address this problem usually rely on image features that are difficult to interpret. Particularly, these image features are subdivided into global and local features, where global features are extracted from the global representation of the image,  while local features are extracted from the objects detected locally in an image. Although, local features extract rich visual information from the image, existing models generate captions in a blackbox manner and humans have difficulty interpreting which local objects the caption is aimed to represent. Hence in this paper, we propose a novel framework for the image captioning with an explicit object (e.g., knowledge graph entity) selection process while still maintaining its end-to-end training ability. The model first explicitly selects which local entities to include in the caption according to a human-interpretable mask, then generate proper captions by attending to selected entities. Experiments conducted on the MSCOCO dataset demonstrate that our method achieves good performance in terms of the caption quality and diversity with a more interpretable generating process than previous counterparts.
\end{abstract}

\section{Introduction}
\label{sec:intro}
Over the past few years, the task of generating descriptions for images (i.e., image
captioning)~\cite{vinyals:2015,anderson:2017} has become popular as it effectively brings together vision and natural language to serve various real-world applications. Most of the existing approaches are efficient in learning a correspondence between image and sequence of words with different techniques that either improve how visual information is captured with attention~\cite{xu:2015,lu:2016,anderson:2017} or language model
interactions~\cite{shen2017estimation}. 

Careful analysis of methods that aim to effectively capture visual information reveal that either utilize global image features or attend to regions for local image features to generate captions. However, this makes it hard to interpret, as they do not select or control objects in an image which may be prominent for caption generation. It is especially important for easy understanding of the caption generation process in case of failures in those systems that cater real-world applications such as autonomous driving, medical imaging and surveillance. Also, observed
previously~\cite{wangobject:2018} that rich entities and their interactions in some kind of a layout can help to better understand image captioning.

\begin{figure*}
    \centering
        \includegraphics[width=0.6\textwidth]{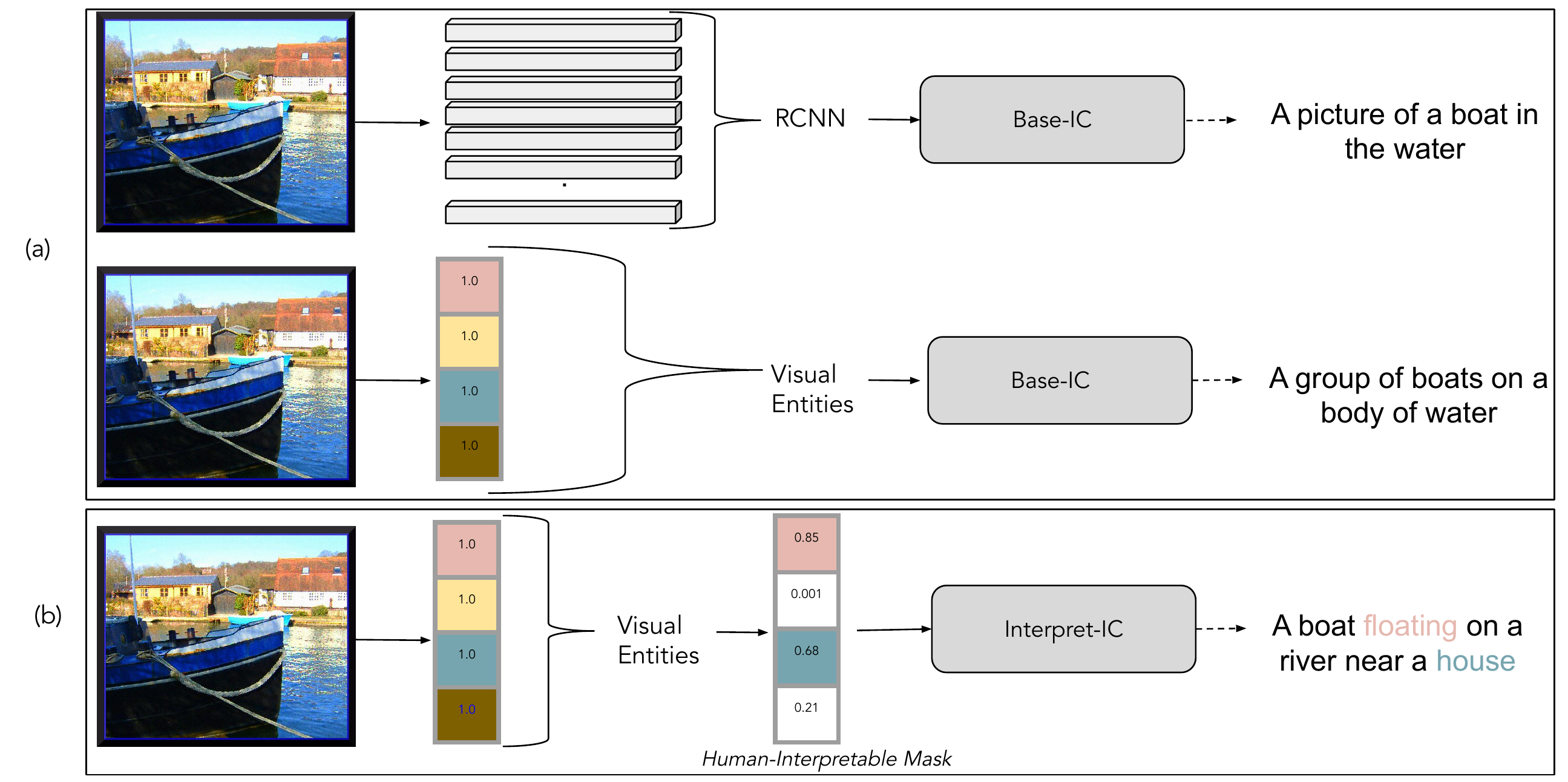}
    \caption{Comparison between two proposed models with different visual features (a) Base-IC (Section~\ref{ssec:baseic}) and (b)  Interpret-IC (Section~\ref{ssec:interpretic}). Interpret-IC has an extra process of highlighting which objects to cover in the generated caption from a shortlist of all detected objects in the image. }\label{fig:sample}
\end{figure*}

Therefore, in this paper, we introduce our interpretable image caption generation model (henceforth, Interpret-IC) to address the limitations of previous approaches as shown in the Figure~\ref{fig:sample}. 
Our proposed approach work with a \textit{human-interpretable mask} which selects the set of local objects observed in an image based on human proposed rules. These rules ensure that only those desirable objects are selected which human wants to observe in the caption. For this to work, the local objects need to 
be represented with semantically enriched labels so that humans can comprehend. As none of the current approaches
provide such local object information. We leveraged relational knowledge provided by the knowledge graph entities to attain semantic labels by building a multi-label image classifier and replace local object visual features with entity distributed representations~\cite{bordes:2011}. We show that these entity labels and its features are superior detected local object features in terms of interpreting knowledge from the image. Very close to our approach by~\cite{cornia:2019}, who considers the decomposition of a sentence into noun chunks and models the relationship between image regions and textual chunks. However, we dynamically select the number of objects prior learning the model. Our main contributions are as follows:

\begin{itemize}
 \item We proposed a novel end-to-end caption model for intepretable image captioning.
 \item We used knowledge graph entities as image labels for grounding visual and factual knowledge.
 \item We show that interpretable image captioning can attain diversity in the captions generated with simple visual object masking. 
 \end{itemize}

\section{Related Work}

In the related work, we explore deep neural network based approaches which generate sentence-level natural language description for images.

\subsection{Diverse Image Captioning}
In the recent years, monolingual image caption generation is explored to
incorporate diversity in the generated captions. Approaches~\cite{ligen:2018} has leveraged 
adversarial training using either generative adversarial networks~\cite{shetty:2017} or variational auto-encoder~\cite{shen2019select}. While,~\cite{vijayakumar:2016} used diverse beam search to decode diverse image captions in English. Approaches were also proposed to describe images from cross-domain~\cite{chen:2017}. However, our goal in this research is to provide better selection procedure for identifying preferable objects in images. Nevertheless, we show that interpretability can also assist diversity.
\subsection{Controllable Image Captioning}
Approach that is closer to intepretable image captioning is a procedure to control local objects in images.~\cite{cornia:2019} used either a sequence or a set of local objects by explicitly grounding them with noun chunks observed in the captions to generate diverse captions. Further, instead of making captions only diverse,~\cite{deshpande:2019} made the captioning more accurate. Our work falls into this space, however understanding the important entities that represent the image and controlling them is what we aim to achieve. 

\section{Interpretable Image Captioning}
\label{}

\subsection{Base-IC Model}
\label{ssec:baseic}
\begin{figure}
    \centering
        \includegraphics[width=0.20\textwidth]{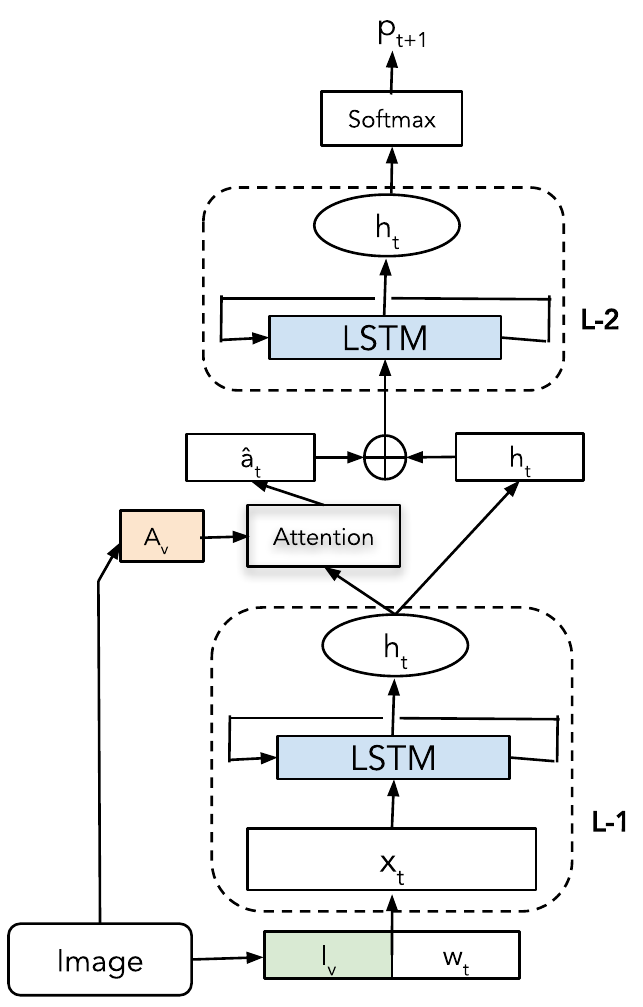}
    \caption{Illustration of Base-IC Model}\label{fig:archibaseic}
\end{figure}

The base image caption model (Base-IC) is built \textit{without} masking. Given an image $I$, its global representation $\boldsymbol{I}_v\in \mathbb{R}^V$ denote the encoding of the full image, while the spatial objects ${a}_v$ = $\{{a}_{v_1},\ldots,{a}_{v_L}\}$ encode local regions of the image provided as $\boldsymbol{a}_{v_j} \in \mathbb{R}^D$ . Similar to previous 
works~\cite{lu:2016,anderson:2017}, our proposed image description model also leverages soft attention mechanism to weigh spatial objects during description generation using the partial output sequence as context. Figure~\ref{fig:archibaseic} illustrates the architecture.

Initially, L-1 of the model receives input from the global visual context provided by $\boldsymbol{I}_v$ and textual sequence, where each word ($\boldsymbol{w}_t \in \mathbb{R}^T$) at time step $t$ in the textual sequence is initialized with the pretrained word embeddings to produce hidden vectors $\boldsymbol{h}_{t}^1 \in \mathbb{R}^{H_1}$. Furthermore, $\boldsymbol{h}_{t}^1$ is used in combination with $\boldsymbol{a}_v$ to compute soft attention. Later, $\boldsymbol{h}_{t}^1$ and attended spatial features are added and provided as input to L-2 for attaining $\boldsymbol{h}_{t}^2 \in \mathcal{R}^{H_2}$. For convenience and to reduce many parameter names, we use $\Theta$ as the reference for the parameters of the LSTM. 

To calculate attended spatial features ($\hat{\boldsymbol{a}}_t$) we leverage $\boldsymbol{a}_v$. Hidden sequences $\boldsymbol{h}_{t}^1$ at 
each time step $t$ is used to generate a normalized attention weight $\boldsymbol{\alpha}_t$ for each of the spatial object features ($\boldsymbol{a}_{v_j}$) given by Equation~\ref{eqn:atj} and Equation~\ref{eqn:etj}.
\begin{equation}
\small
\label{eqn:atj}
 \boldsymbol{\alpha}_{tj}= \frac{exp(\boldsymbol{e}_{tj})}{\sum^L_{k=1} exp(\boldsymbol{e}_{tk})}
\end{equation}
\begin{equation}
\small
\label{eqn:etj}
 \boldsymbol{e}_{tj} = \text{tanh}(W_{ae}\boldsymbol{a}_{v_j}+ W_{he}\boldsymbol{h}_{t}^1)
\end{equation}
where $L$ represent the cardinality of set $a_{v}$. $W_{ae} \in \mathbb{R}^{M \times D}$ and $W_{he} \in \mathbb{R}^{M \times H_1}$ are learnable parameters. Further, $\hat{\boldsymbol{a}}_t$ is calculated with Equation~\ref{eqn:ahat} and is used as input along with $\boldsymbol{h}_{t}^1$ to the L-2 at every time step $t$.
\begin{equation}
\small
\label{eqn:ahat}
 \hat{\boldsymbol{a}}_t= \sum_{j=1}^L \boldsymbol{\alpha}_{tj}a_{v_j}
\end{equation}
The final Base-IC using $\boldsymbol{w}_{t}$ and $\boldsymbol{I}_v$ as input to L-1 is given by Equation~\ref{eqn:input} and $\boldsymbol{h}_{t}^1$ is given by Equation~\ref{eqn:gvcoh1}. Further, $\hat{\boldsymbol{a}}_t$ and $\boldsymbol{h}_{t}^1$ are added using Equation~\ref{eqn:xprimet} to provide as input for L-2 for generating $\boldsymbol{h}_{t}^2$ as given by Equation~\ref{eqn:gvcih2}. It is then used to predict next words in the sequence as given in the Equation~\ref{eqn:modhid}. 
\begin{equation}
\small
\label{eqn:input}
      \boldsymbol{x}_t = \boldsymbol{I}_v \oplus \boldsymbol{w}_t
\end{equation}
\begin{equation}
\small
\label{eqn:gvcoh1}
    \boldsymbol{h}_{t}^1 = \text{L-1}(\boldsymbol{x}_t,\boldsymbol{h}_{t-1}^1;\Theta)
\end{equation}

\begin{equation}
 \small
 \label{eqn:xprimet}
       \boldsymbol{x}'_{t} =  \hat{\boldsymbol{a}}_t + \boldsymbol{h}_{t}^1
 \end{equation}
\begin{equation}
 \small
 \label{eqn:gvcih2}
     \boldsymbol{h}_{t}^2 = \text{L-2}(\boldsymbol{x}'_t,\boldsymbol{h}_{t-1}^2;\Theta)
 \end{equation}
\begin{equation}
\small
\label{eqn:modhid}
    p_{t+1} = \text{softmax}(W_{vocab}\boldsymbol{h}_{t}^2)
\end{equation}
where $W_{vocab} \in \mathbb{R}^{vocab \times (V+H_2)}$ , $\oplus$ represents concatenation and $vocab$ refers to vocabulary of the caption dataset.

\subsection{Interpret-IC Model}
\label{ssec:interpretic}
Main aim of the \textit{Interpret-IC} model is to select objects present in the spatial objects set ${a}_v$ with human-interpretable masking. This is in contrast with earlier 
approaches~\cite{xu:2015,anderson:2017}, who decoded the caption by attending to spatial objects only by ranking them according to their importance at each time step. Also, these approaches provide no control for 
humans to select their desirable objects. It clearly sets expectation from \textit{Interpret-IC} model that the selected objects should provide more prominence in caption generation by discarding those objects that are not selected. 

Hence, we introduce \textit{masked attention} to select those objects that human wants to see in the generated captions. To achieve it, we leverage ground truth \textit{mask} i.e., $\text{mask}_{gt}$ where each object in the ${a}_v$ is masked with a binary parameter $\beta_1,\beta_2,\ldots,\beta_n$. We set $\beta_i=1$ if selected and 0 otherwise. Also, $\beta_i$ is assumed to be independent from each other and is sampled from a bernoulli distribution. Prediction mask i.e., $\text{mask}_{pred}$ is estimated during training with a multi-layer perceptron (MLP). 

\begin{figure}
    \centering
        \includegraphics[width=0.35\textwidth]{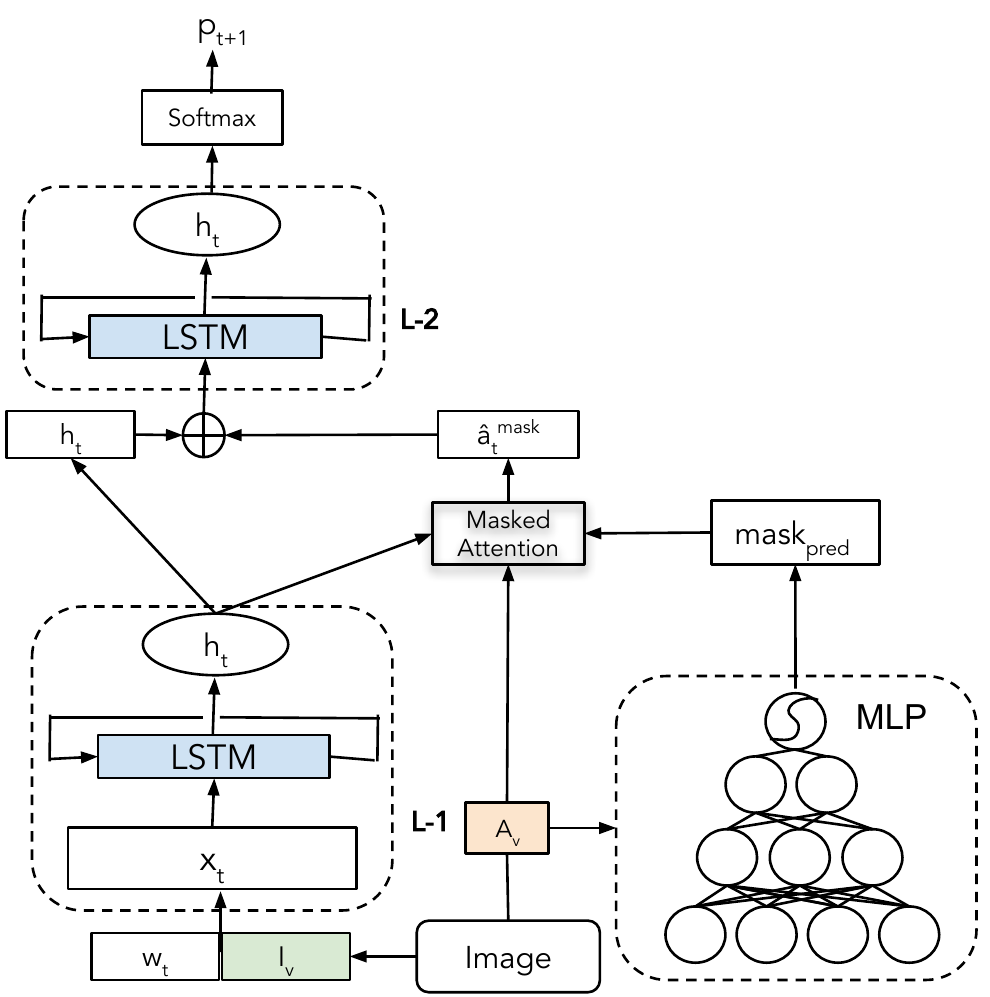}
    \caption{Illustration of Interpret-IC Model}\label{fig:archiinterpret}
\end{figure}

Further, attention weights computed in the Equation~\ref{eqn:atj} is modified with the estimated $\text{mask}_{pred}$  as shown in Equation~\ref{eqn:modatj}.
\begin{equation}
 \label{eqn:modatj}
 \small
  \boldsymbol{\alpha}_{tj}^{mask}= \frac{exp(\boldsymbol{e}_{tj})\text{mask}_{pred}}{\sum^L_{k=1} exp(\boldsymbol{e}_{tk})\text{mask}_{pred}}
\end{equation}
It is then used to calculate $\hat{\boldsymbol{a}}^{mask}_t$ given by Equation~\ref{eqn:amaxhat}, which is further used as input along with $\boldsymbol{h}_{t}^1$ to the L-2 at every time step $t$. Figure~\ref{fig:archiinterpret} illustrates the overall architecture.
\begin{equation}
\small
\label{eqn:amaxhat}
 \hat{\boldsymbol{a}}^{mask}_t= \sum_{j=1}^L \boldsymbol{\alpha}^{mask}_{tj}a_{v_j}
\end{equation}
Note that our selection strategy is very different from~\cite{cornia:2019}, who control spatial objects using the fixed noun-chunks extracted from captions which are not available during testing phase. While, we use human designed rules to change our mask, so that we control the mask as we aim to use it.    

\subsection{Ground Truth Mask Selection}
\label{ssec:masksel}
In the \textit{Interpret-IC} model, $\text{mask}_{pred}$ needs to be optimized during training phase closer to the ground truth binary mask i.e., $\text{mask}_{gt}$ such that it can be utilized during the testing phase. However, first we need to create such $\text{mask}_{gt}$  based on \textit{human-interpretable rules} to influence the caption generation process. 

There can be several ways to create $\text{mask}_{gt}$ by changing the rules. In this paper, we apply \textit{visual entities} to caption \textit{noun} matching approach to build the $\text{mask}_{gt}$. Our \textit{rule} here states that for each noun identified\footnote{\url{https://spacy.io/}} in the caption, we need to find the closest visual entity by computing cosine distance between the noun and visual entity vectors attained using pretrained fastText\footnote{\url{https://fasttext.cc/}} vectors. For all nouns identified, closest visual entities are set to 1, while rest are set to 0. This rule ensures that the nouns observed in the caption representing some kind of objects present in images have to be given higher preference during caption generation. While, rest of the visual entities (e.g., actions) are put on back burner.  Algorithm~\ref{alg:selcti} presents the overview of selection process.

\begin{algorithm}[h]

\SetAlgoLined
    \KwInput{Nouns (N), Visual Entities (VE), fastText Embeddings (FTE)}
    \KwOutput{$\text{mask}_{gt}$ for each caption}
\SetKwProg{generate}{Function \emph{$\text{mask}_{gt}$ Selection}}{}{end}
Initialize $\text{N}_{emb}$ = FTE(N) \;
Initialize $\text{VE}_{emb}$ = FTE(VE) \;
Initialize $\text{Image}_{velist}$ as $\text{I}_{velist}$\;
Initialize $\text{Caption}_{list}$ as $\text{C}_{list}$\;
\generate{}{
     \For{C,VE in $\text{C}_{list}$,$\text{I}_{velist}$}{
      Extract N from caption \;
      Initialize $\text{mask}_{gt}$ = zeros[len(VE)] \;
     \For{n in N}
     {
     \If{n not EMPTY}{
     dist = CosineDistance($\text{n}_{emb}$,$\text{VE}_{emb}$);
     $\text{close}_{index}$ = $\argmin(\text{dist})$\;
     $\text{mask}_{gt}[\text{close}_{index}]$ = 1\;
     }
    }
    \Return $\text{mask}_{gt}$ \;
    }
}
\caption{\label{alg:selcti} $\text{mask}_{gt}$ selection process}
\end{algorithm}
 \section{Training and Inference}
\paragraph{Base-IC}
The parameters ($\theta$) of the \textit{Base-IC} model are trained for optimizing the cost function ($\mathcal{C}$) to minimize the sentence-level categorical cross-entropy loss by finding negative log likelihood of the appropriate ground truth word ($y_{t}^*$) at each time step $t$ as shown in Equation~\ref{eqn:baseloss}. Here, we leverage teacher forcing~\cite{sutskever:2014}, where ground truth ($y_{t}^*$) is fed to next step in the layer L-1, instead of the predicted word in previous step. 
\begin{equation}
\label{eqn:baseloss}
\small
\mathcal{C}(\theta) = -\sum_{t=0}^{T^{(n)}} \text{log} p_{\theta}(y_{t}^{*})
\end{equation}
The $T^{(n)}$ represents the length of the sentence at $n$-th training sample. During inference, we leverage beam search with beam size is set to 5 in our experiments.
\paragraph{Interpret-IC}
Similar to \textit{Base-IC} model, parameters ($\theta^{'}$) of the \textit{Interpret-IC} are trained for optimizing the cost function ($\mathcal{C}^{'}$) which minimizes both the sentence-level categorical cross-entropy loss along with binary cross-entropy loss that approximate ($\text{mask}_{pred}$) closer to the ground truth mask ($\text{mask}_{gt}$) as shown in Equation~\ref{eqn:inticloss}.

\begin{equation}
\label{eqn:inticloss}
\begin{split}
\small
\mathcal{C}^{'}(\theta^{'}) = &-\Bigg(\Big(\sum_{t=0}^{T^{(n)}} \text{log} p_{\theta}(y_{t}^{*})\Big) + \text{mask}_{gt}log(\text{mask}_{pred}) \\&+ (1-\text{mask}_{gt})log(1-\text{mask}_{pred})\Bigg)    
\end{split}
\end{equation}
During inference, similar to \textit{Base-IC} model, we leverage beam search by setting beam size to 5 in our experiments.
\section{Evaluation Setup}
\label{sec:eval}
\paragraph{Datasets} For experimental evaluation, we use MSCOCO dataset with splits of~\cite{karpathy:2015}. Table~\ref{statistics-datasets} summarizes the training, validation and test splits. 
\begin{table}[!ht]
\centering
\small
\begin{tabular}{lc}
\toprule
Mean Sentence-Length &  11.3 \\
Vocabulary & 9989\\
Sentences & 5 \\
Training & 113287 \\
Validation & 5000\\
Test & 5000 \\
\bottomrule
\end{tabular}
\caption{\label{statistics-datasets}Statistics of the MSCOCO dataset}
\end{table}
\subsection{Implementation}
\label{ssec:impl}

\paragraph{Local and Global Image Features}

Spatial object ($\boldsymbol{a}_v$) features are extracted in two different ways.
\begin{itemize}
 \item Faster R-CNN~\cite{ren:2015} in conjunction with the ResNet-101~\cite{he:2016} trained on visual genome data by~\cite{anderson:2017} is used to extract top 36 local object features ($\boldsymbol{a}_{v_j}$) of dimension 2048. There are pure visual features and we refer to this set as Obj$\rightarrow$RCNN.
 \item Since, Obj$\rightarrow$RCNN represent pure visual features without label information. Following~\cite{mogadala:2018,mogadaladiscover:2018}, we extracted semantically enriched labels denoting entities from captions aligned to an image in training set of MSCOCO with a knowledge graph annotation tool such as DBpedia spotlight\footnote{\url{https://github.com/dbpedia-spotlight/}}. In total, 812 unique human-interpretable already disambiguated labels are extracted. Further, a multi-label image classifier is trained with sigmoid cross-entropy loss by fine-tuning VGG-16~\cite{simonyan:2014} pre-trained on the training part of the ILSVRC12 with training images in MSCOCO. After training, we use the classifier to acquire Top-15 entity labels for each image present in the training, validation and testing set of MSCOCO. Now, to use entity labels similar to Obj$\rightarrow$RCNN features. We use knowledge graph embeddings~\cite{ristoski:2016} and generate 500 dimensional vectors\footnote{Please note that these embeddings are different from fastText Vectors used to build $\text{mask}_{gt}$. These embeddings are analogous to pure visual features, however, learned from knowledge graph structure.} for each entity-label. We refer to this set as Obj$\rightarrow$VisualEntity.
 \item The global visual features ($\boldsymbol{I}_v$) of dimension 2048 is extracted using the average pooling of Obj$\rightarrow$RCNN features.
\end{itemize}

\paragraph{Caption Model}

Both \textit{Base-IC} and \textit{Interpret-IC} models are built by initializing the model with input ($\boldsymbol{w}_t$) word embeddings pretrained using Glove~\cite{pennington:2014} on the MSCOCO training captions corpora. The dimensions of the hidden units $\boldsymbol{h}_t^{1},\boldsymbol{h}_t^{2}$ in \textbf{L-1} and \textbf{L-2} of models are set to 512. Also, the hidden units of shared 
layer $\boldsymbol{h}_t^{(s)}$ are set to 512. All models are then trained with Adam optimizer with gradient clipping having maximum norm of 1.0 and mini-batch size of 50 for 25 epochs. Initially, the learning is set to 0.001 and is reduced by a factor of 10 if there is no improvement in the validation loss for 3 continuous epochs.

\paragraph{Evaluation Measures}

We first evaluate the generated captions based on correctness which guarantee the generation quality based on standard captioning metrics. 
Further, we check if our proposed model with human-interpretable masking can generate diverse and interesting captions. For this, we leverage earlier proposed~\cite{shetty:2017,deshpande:2019} metrics such as vocabulary size and novel caption with best (i.e., Top-1) generated caption.  Vocabulary Size (VS) find unique words in generated captions and Novel captions (NC) identify the percentage of generated captions that are not seen in the training set.
\begin{table*}[htbp]
\small
  \centering
  \begin{tabular}{lccccc}
    \toprule
      & \multicolumn{5}{c}{Cross-Entropy Loss} \\
    \cmidrule(lr){2-6} 
    &\multicolumn{1}{c}{BLEU-4} & \multicolumn{1}{c}{METEOR}  & \multicolumn{1}{c}{ROUGE-L} & \multicolumn{1}{c}{CIDEr} &  \multicolumn{1}{c}{SPICE}  \\
    \cmidrule(lr){2-6} 
    {Model} &  &  & & & \\
    \midrule
    Adv-bs~\cite{shetty:2017} &  - & 23.9 & - & - & 16.7 \\
    CNN+CNN~\cite{wangcnnplus:2018} &  26.7 & 23.4 & 51.0 & 84.4 & - \\
    Convolutional-IC~\cite{aneja:2018} &  31.6 & \textbf{25.0} & 53.1 & 95.2 & 17.9 \\
    POS+Joint~\cite{deshpande:2019} & - & 24.7 & - & - & 18.0 \\
    \midrule
    Base-IC & & &  & &\\
    +Obj$\rightarrow$RCNN  & 31.8 & 24.9 & 52.9 &  96.7 & \textbf{18.2}   \\
    +Obj$\rightarrow$VisualEntity  & 32.1 & 24.8 & 53.6 & 96.9 & 18.0 \\
    \cmidrule(lr){2-6}
    Interpret-IC & & &  & &\\
    +Obj$\rightarrow$VisualEntity & \textbf{32.4} & 24.9 & \textbf{53.7} & \textbf{97.8} & 18.1 \\
    \bottomrule
  \end{tabular}
  \caption{\label{standardcocoresults} Results achieved with our models in comparison with baseline approaches.}
\end{table*}
\begin{figure*}
    \centering
        \includegraphics[width=0.85\textwidth]{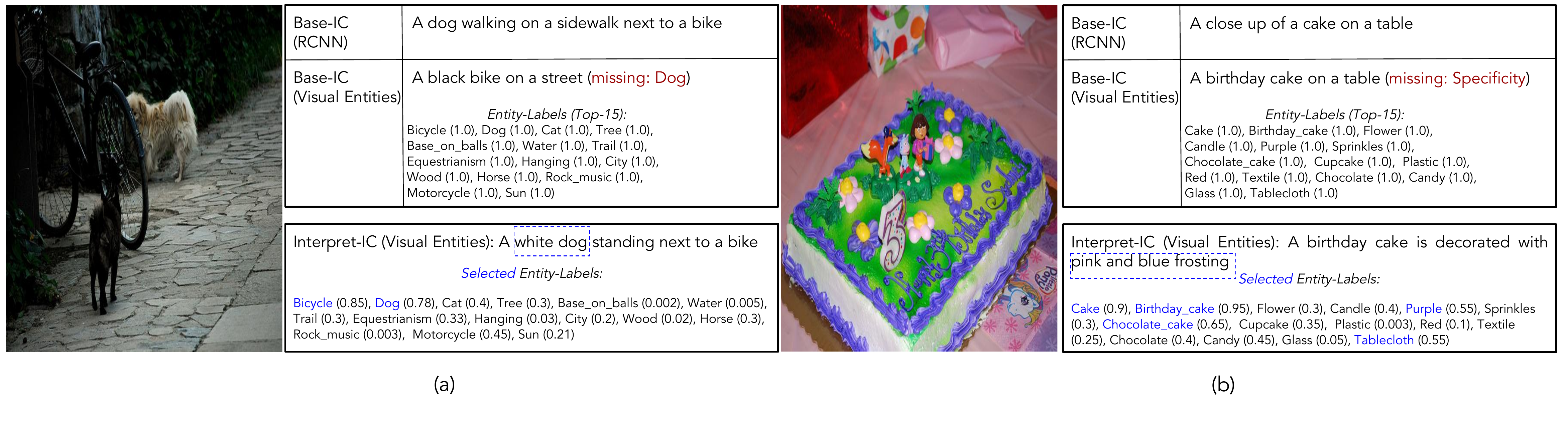}
    \caption{\textbf{Caption Coverage Example} (Entities with $\text{mask}_{pred} > 0.5$ are highlighted in blue): (a) Missing local object (Dog) in the caption generated by \textit{Base-IC}, while ``White Dog'' is included by \textit{Interpret-IC} providing better coverage. (b) Missing details about the birthday cake, \textit{Interpret-IC} generated better and interesting caption by highlighting objects that need to be focused on.}\label{fig:cmsamp}
\end{figure*}
\begin{figure*}
    \centering
        \includegraphics[width=0.85\textwidth]{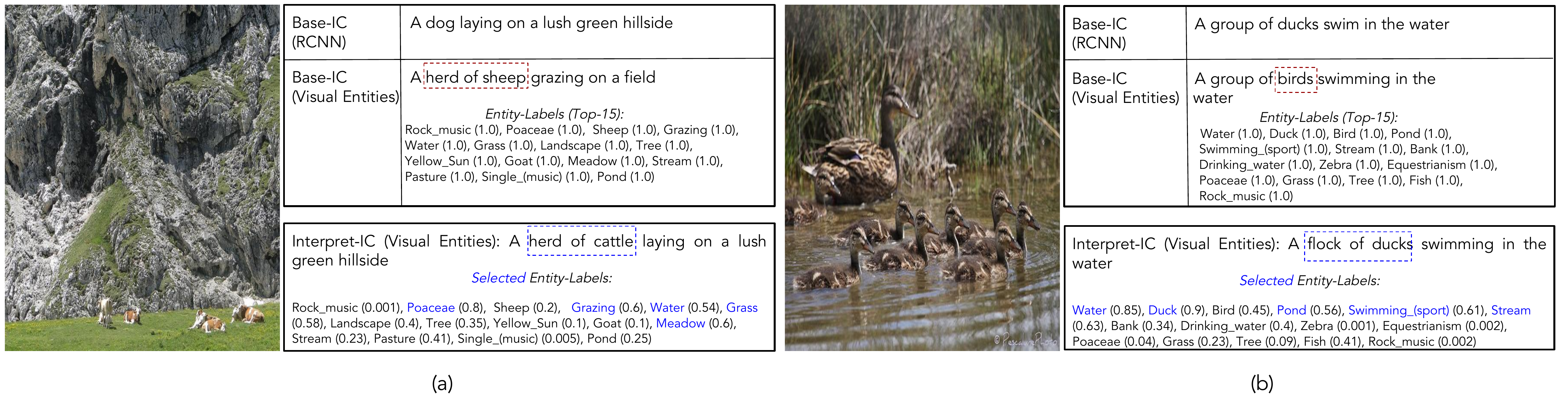}
    \caption{\textbf{Caption Correction Example} (Entities with $\text{mask}_{pred} > 0.5$ are highlighted in blue): (a) \textit{Base-IC} generate a caption by including wrong objects i.e., sheep, while ``cattle'' is included by \textit{Interpret-IC} because a lower weight (0.2) is assigned to ``sheep'' hence filters out the wrongly detection. (b) Although \textit{Base-IC} covers the correct object (Birds), it is too general and fails to provide more informative caption. \textit{Interpret-IC} replaces it with the exact object by giving a large weight to emphasize the detected entity ``duck".}\label{fig:ccsamp}
\end{figure*}
\begin{figure*}
    \centering
    \begin{subfigure}[b]{0.35\textwidth}
        \includegraphics[width=\textwidth]{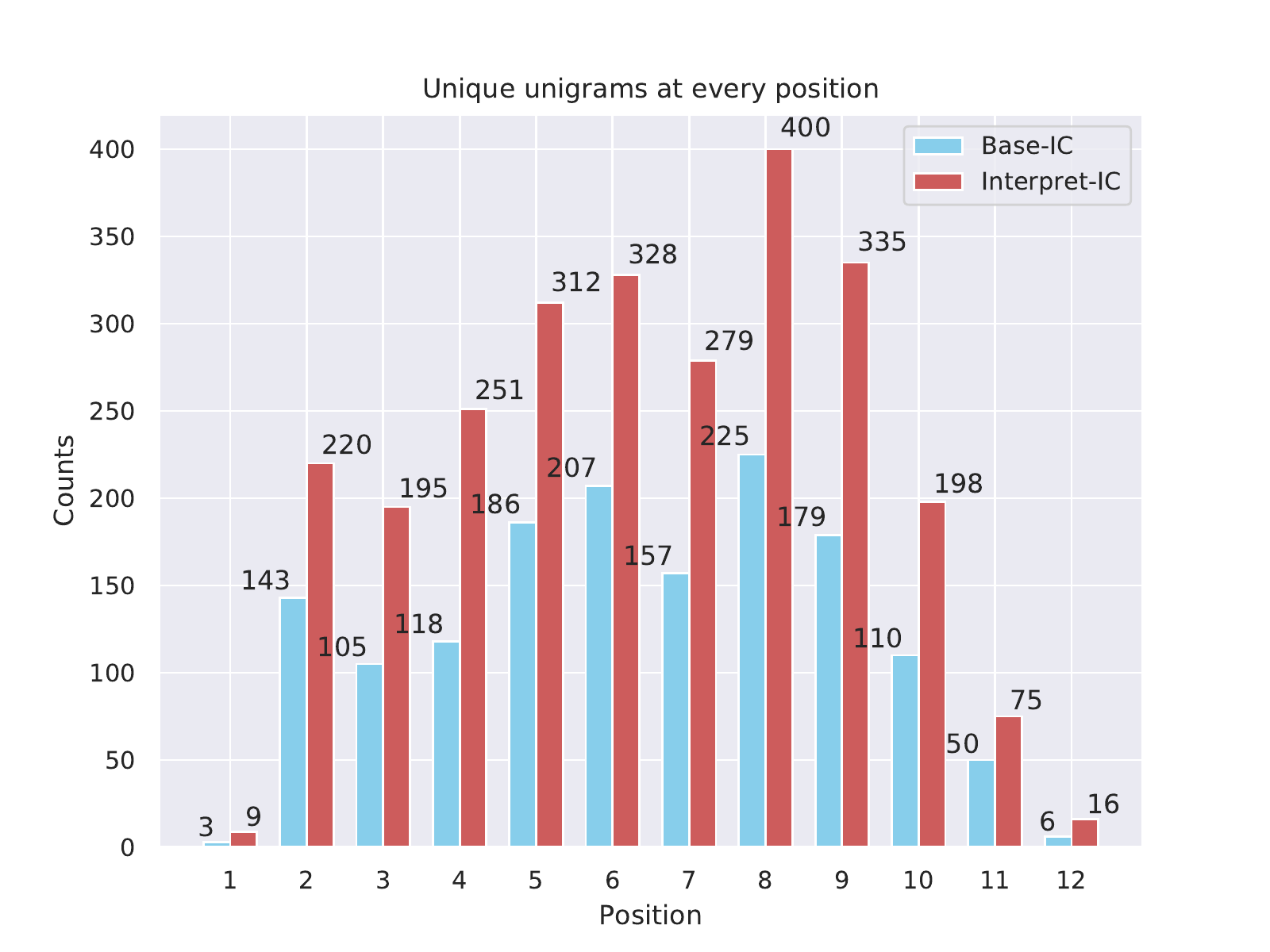}
        \caption{Unigrams}
        \label{fig:unigrams}
    \end{subfigure}
    ~ 
    \begin{subfigure}[b]{0.35\textwidth}
        \includegraphics[width=\textwidth]{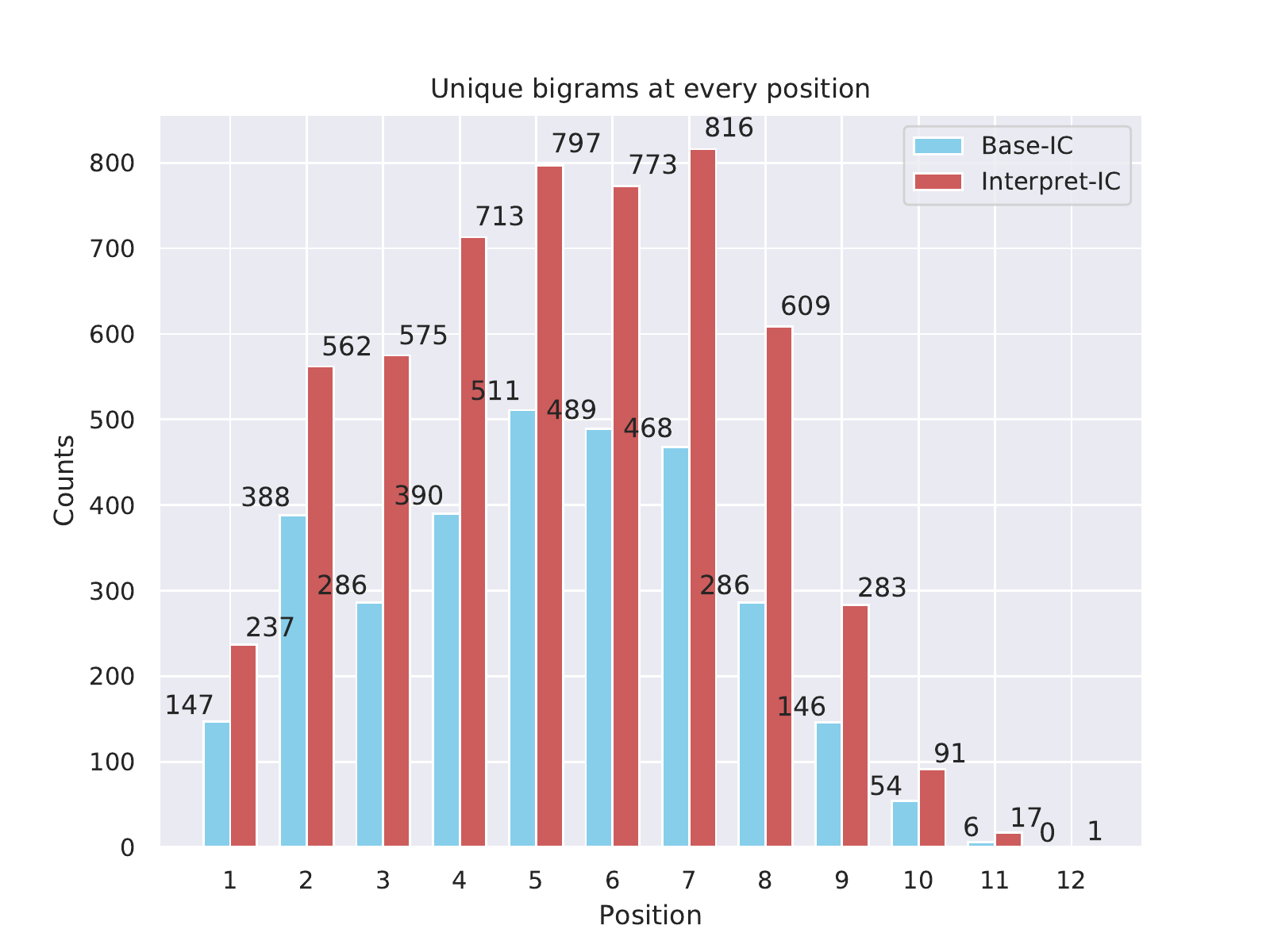}
        \caption{Bigrams}
        \label{fig:bigrams}
    \end{subfigure}
    \caption{\label{fig:unibi} Plot of starting unique unigrams and bigrams observed in the generated caption. }
\end{figure*}
\section{Results}
\label{sec:evalres}
\subsection{Quantitative Results}
We compared our proposed \textit{Base-IC} and \textit{Interpret-IC} along with other recent baselines. Table~\ref{standardcocoresults} shows the results obtained. It can be observed that the \textit{Interpret-IC} model was able improve over recent approaches by allowing better control over the caption generation process.

\subsection{Qualitative Results}
To understand the contribution made by \textit{human-interpretable mask} to caption generation. We explored qualitatively the captions generated by both \textit{Base-IC} and \textit{Interpret-IC} models with visual entities from two different perspectives. First, we observed the quality of the predicted mask in selecting required visual entities for better coverage. Second, we checked if \textit{Interpret-IC} model could overcome or correct mistakes made by the \textit{Base-IC} model. In the following, we discuss each of these cases briefly by showing some examples.

\paragraph{Caption Coverage}
We use visual entities such that they represent local objects in images to be incorporated them in the caption. However, this cannot be simply achieved with a \textit{Base-IC} model. As seen in Figure~\ref{fig:cmsamp}, the \textit{Interpret-IC} model which weighs each of these objects differently based on the predicted mask, when compared with the \textit{Base-IC} model giving equal importance to each of them. Although the Base-IC model generated partially relevant caption, masking has shown to improve coverage of local objects in the image. The selector is able to assign higher scores to prominent objects in the image which increases the probability of covering them in the generated caption.

\paragraph{Caption Correction}
We also observe that, apart from providing better coverage of visual entities in the generated captions. Masking also plays a prominent role in the caption correction. That is, as seen in the Figure~\ref{fig:ccsamp}, although the \textit{Base-IC} model generated a partially relevant caption, \textit{Interpret-IC} generated the most accurate caption with effective selection of relevant visual entities. The selector is expected to assign lower scores to inappropriate (bird in Figure~\ref{fig:ccsamp}b) or wrongly detected objects~(sheep in Figure~\ref{fig:ccsamp}a) thus encouraging the decoder to attend to more plausible entities.
\subsection{Diversity}
Although our aim is not to achieve diverse captions, to comprehend whether our proposed \textit{Base-IC} and \textit{Interpret-IC} models generate best (i.e., Top-1) diverse and interesting caption. We compared our models with other diverse caption generation baselines that compare best generated caption using diversity measures described earlier. Table~\ref{diversitycocoresults} shows the results attained , where (NC) is Top-1 generated caption with Base-bs~\cite{shetty:2017} and Adv-bs~\cite{shetty:2017}. We observe that, our \textit{Interpret-IC} model cannot exceed scores of the baseline trained to generate diverse captions in an adversarial setting (i.e., Adv-bs). However, with less effort and simple masking we could see a significant jump on the standard caption model (i.e., Base-bs).

\begin{table}[!ht]
\small
  \centering
  \begin{tabular}{lcccc}
    \toprule
    &\multicolumn{1}{c}{Base-bs} & \multicolumn{1}{c}{Adv-bs} & \multicolumn{1}{c}{Base-IC} & \multicolumn{1}{c}{Interpret-IC}    \\
    \cmidrule(lr){2-5} 
    {Metrics} &  & & & \\
    \midrule
    VS & 756 & \textbf{1508} & 443 & 862 \\
    NC & 34.18 & \textbf{68.62} & 36.23 & 51.54 \\
    \bottomrule
  \end{tabular}
  \caption{\label{diversitycocoresults} Diversity: Comparison of vocab size (VS) and novel captions. Base-IC and Interpret-IC use +Obj $\rightarrow$ VisualEntity features.}
\end{table}
Also, in Figure~\ref{fig:unibi}, we plot unique unigrams and bigrams predicted at every word position. The plot shows that the \textit{Interpret-IC} have higher unique unigrams at different word positions and is consistently higher for the bigrams when compared against \textit{Base-IC} with visual entities as features. This supports our hypothesis that \textit{Interpret-IC} can produce more diverse captions as it can alter caption generation process.

\section{Conclusion and Future Work}
In this paper, we aimed to address the problem of interpretable image captioning by leveraging knowledge graph entity features. Initially, we obtained local objects as visual entities in the image by grounding knowledge graph entities. Further, the human-interpretable masking rules are developed to select those visual entities for generating desirable captions. Experimental results show that interpretability in caption generation can help to alter caption generation process hence allowing control and selection. In Future, we aim to improve caption generation process by trying different masks and better sampling.

\section{Acknowledgements}
\label{sec:ack}
Aditya Mogadala was supported by the German Research Foundation (DFG) as a part of - Project-ID 232722074 - SFB1102.

\bibliographystyle{named}
\bibliography{ijcai20}
\end{document}